\documentclass[letterpaper, 10 pt, conference]{ieeeconf}  

\IEEEoverridecommandlockouts  

\usepackage{graphicx} 
\usepackage{amsmath}
\usepackage{float}
\usepackage{subfloat}
\usepackage{caption}
\usepackage{subcaption}
\usepackage[font=small,labelfont=bf]{caption}
\usepackage[compact]{titlesec}
\usepackage{url}

\usepackage{bm}
\usepackage[procnumbered, ruled, linesnumbered]{algorithm2e}
\usepackage{etoolbox}

\usepackage{multirow}

\usepackage{balance}

\usepackage[usenames,dvipsnames]{color}
\usepackage[normalem]{ulem}

\makeatletter
\AtBeginEnvironment{procedure}{\let\c@algocf\c@procedure}
\makeatother
\usepackage{amsmath}
\usepackage{amssymb}

\makeatletter
\newcommand{\removelatexerror}{\let\@latex@error\@gobble}
\makeatother
\let\oldnl\nl
\newcommand{\nonl}{\renewcommand{\nl}{\let\nl\oldnl}}

\renewcommand\dbltopfraction{0.99}
\usepackage{soul}
\definecolor{aqua}{rgb}{0.0, 1.0, 1.0}
\definecolor{bittersweet}{rgb}{1.0, 0.44, 0.37}
\definecolor{chartreuse}{rgb}{0.87, 1.0, 0.0}

\def\bfb{{\mbox{\boldmath $b$}}}

\def\bfe{{\mbox{\boldmath $e$}}}

\def\bfq{{\mbox{\boldmath $q$}}}

\def\bfs{{\mbox{\boldmath $s$}}}

\def\bfF{{\mbox{\boldmath $F$}}}

\def\bfH{{\mbox{\boldmath $H$}}}
\def\bfI{{\mbox{\boldmath $I$}}}
\def\bfJ{{\mbox{\boldmath $J$}}}

\def\bfW{{\mbox{\boldmath $W$}}}

\def\ExampleStandUp{\centering\includegraphics[width=0.85\columnwidth]{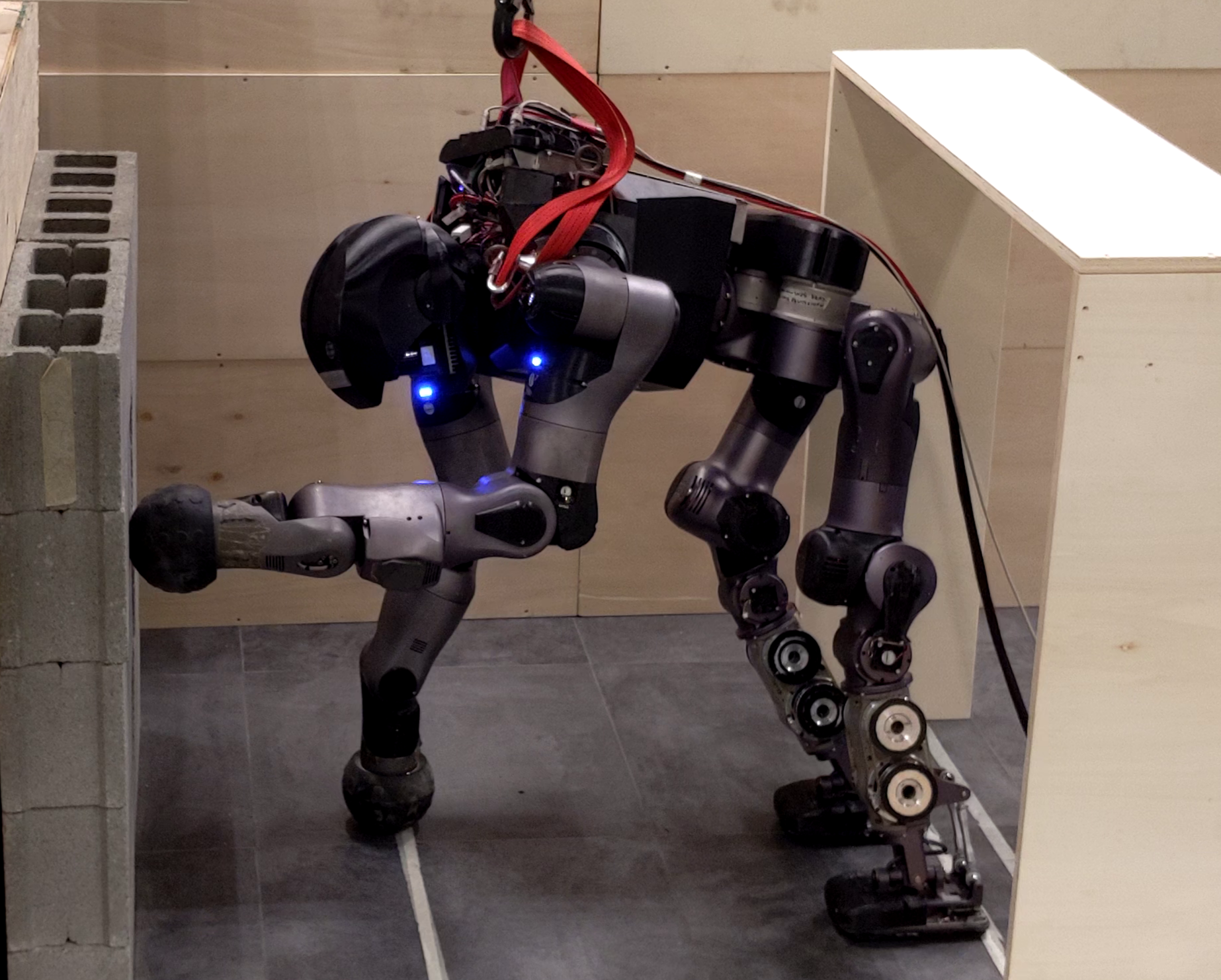}}
\def\Scheme{\centerline{\includegraphics[width=0.85\textwidth]{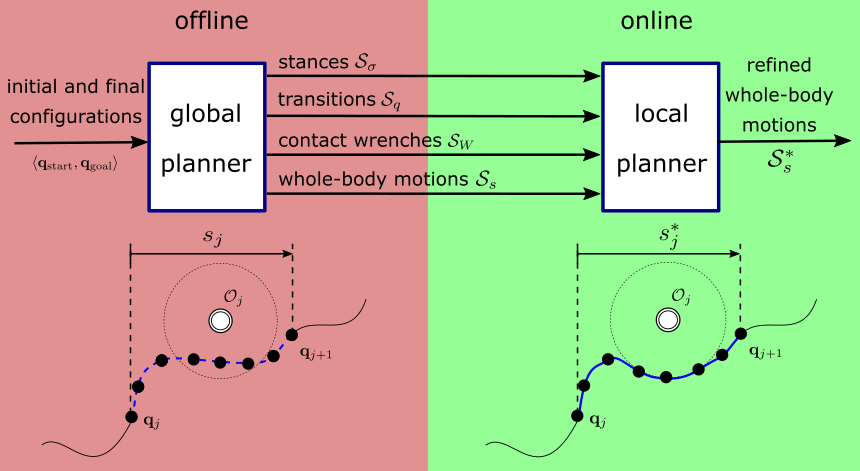}}}

\def\OfflineTasks{\centering\includegraphics[width=1.0\columnwidth]{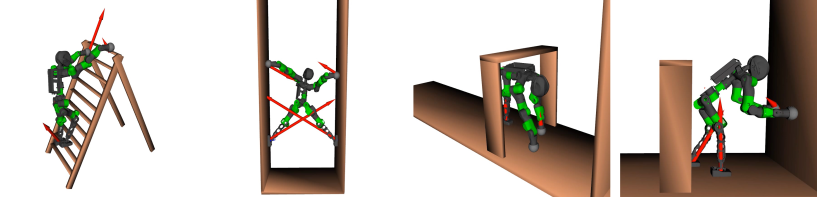}}
\def\COMAN{\centerline{\includegraphics[width=1.0\textwidth]{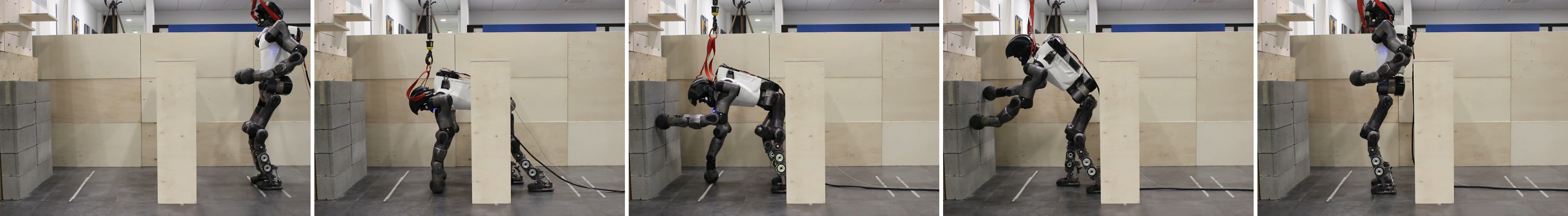}}}

\title{\LARGE \bf
Loco-Manipulation Planning for Legged Robots: \\ Offline and Online Strategies
}

\author{Luca Rossini$^{1,2}$, Paolo Ferrari$^{3}$, Francesco Ruscelli$^{1}$, \\ Arturo Laurenzi$^{1}$, Nikos G. Tsagarakis$^{1}$, Enrico Mingo Hoffman$^4$
	\thanks{$^1$ Humanoid and Human Centered Mechatronics (HHCM) Lab, Istituto Italiano di Tecnologia (IIT), Via Morego 30, 16163 Genova, Italy. 
	E-mail: {\tt\small name.surname@iit.it}}
	\thanks{$^2$ Dipartimento di Informatica, Bioingegneria, Robotica e Ingegneria dei Sistemi (DIBRIS), Universit\`{a} di Genova, Via Opera Pia 13, 16145 Genova, Italy.}
	\thanks{$^3$ Dipartimento di Ingegneria Informatica, Automatica e Gestionale (DIAG), Sapienza Universit\`{a} di Roma, Via Ariosto 25, 00185 Roma, Italy.}
	\thanks{$^4$ PAL Robotics, Carrer de Pujades, 77-79, 08005 Barcelona, Spain.}
}

\begin{document}

\maketitle

\begin{abstract}
The deployment of robots within realistic environments requires the capability to plan and refine the loco-manipulation trajectories on the fly to avoid unexpected interactions with a dynamic environment. 
This extended abstract provides a pipeline to offline plan a configuration space global trajectory based on a randomized strategy, and to online locally refine it depending on any change of the dynamic environment and the robot state.
The offline planner directly plans in the contact space, and additionally seeks for whole-body feasible configurations compliant with the sampled contact states.
The planned trajectory, made by a discrete set of contacts and configurations, can be seen as a graph and it can be online refined during the execution of the global trajectory.
The online refinement is carried out by a graph optimization planner exploiting visual information. It locally acts on the global initial plan to account for possible changes in the environment.
While the offline planner is a concluded work, tested on the humanoid COMAN+, the online local planner is still a work-in-progress which has been tested on a reduced model of the CENTAURO robot to avoid dynamic and static obstacles interfering with a wheeled motion task.
Both the COMAN+ and the CENTAURO robots have been designed at the Italian Institute of Technology (IIT).
\par \textit{Paper Type} -- Recent Work \cite{ijrr:offline}\cite{iros:online}
\end{abstract}

\section{Introduction}
\begin{figure}
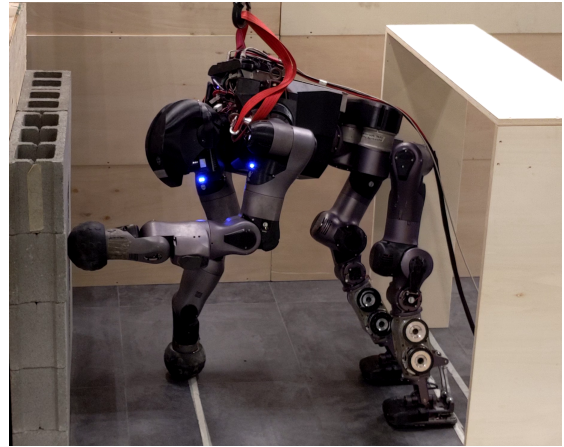

\ExampleStandUp
\caption{The COMAN+ huamnoid robot standing up taking advantage of a vertical wall placed in front of it.}
\label{fig:example_stand_up}
\end{figure}
Humanoids have the potential to accomplish tasks that requires moving in complex and confined spaces by braking and establishing multiple contacts with the environment. 
Examples of these \emph{multi-contact loco-manipulation tasks} include crawling, climbing a ladder and standing up using vertical wall (Fig.~\ref{fig:example_stand_up}).
Additionally, introducing collaborative robots alongside human beings in a domestic or working environment requires the capability of the robot to detect any change in the surrounding environment and react to that on the fly while executing a task.
\begin{figure*}
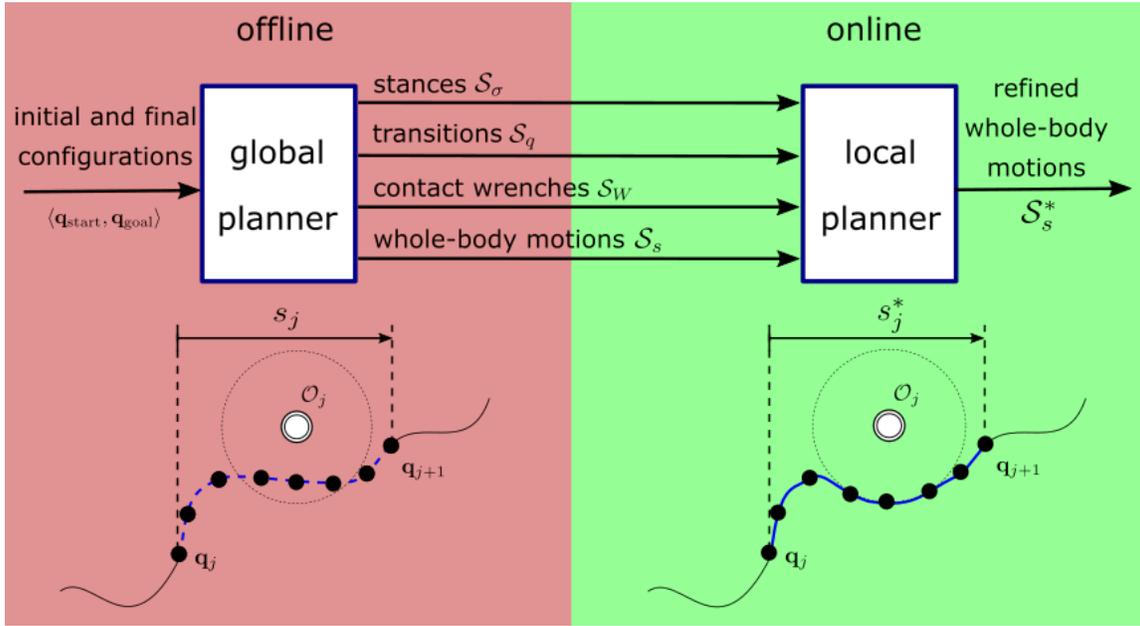

    \centering
    \Scheme
    \caption{Combination of global and local planning. The global planner works offline and provides an initial guess to the online local planner (blue dashed line on the left). While executing the trajectory, the local planner refines it to recover any infeasibility, generating a new feasible motion depending on the environment state. $\mathcal{O}_j $ is a generic obstacle interfering with the global plan computed offline.}
    \label{fig:OnlinePlanner}
\end{figure*}
\par
To effectively fulfill these tasks, the robot must be able to autonomously decide and execute appropriate motion that respect several crucial constraints such as collision avoidance, balance and kinematic/dynamic limitations.
This means addressing a complete Multi-Contact Planning and Control (MCPC) problem. In this extended abstract, we focus specifically on the planning aspects of this problem.

Sample-based global planners revealed to be very efficient over the last decades, and they can easily be combined with perception algorithms avoiding any simplification of the environment~\cite{moveit!}.
In the case of dynamic environment, one possibility would be to continuously re-plan the global trajectory every time a difference in the environment is perceived.
However, sample-based planners revealed to be strongly influenced by the complexity of the robot, and its compuational cost increases exponentially depending on the number of deegres of freedom (DoFs), such as for legged robots.
As a consequence of their massive computation cost, these methodologies are prevented to be used in real-time application (a.k.a. \emph{curse of dimensionality}).
On the other side, optimization-based motion planners can be employed in real-time but they can easily stuck in local minima getting impossible to find an optimal trajectory.
\par
A possible solution is to couple a sample-based planner, generating offline a global initial trajectory, with a optimization-based motion planner exploiting visual information that acts locally while the robot executes the global trajectory (see Fig. \ref{fig:OnlinePlanner}).
In this way, the local planner is globally guided by the global planner avoiding any local minima, while guaranteeing a computational efficiency that allows an online implementation during the execution of the assigned task.
These kind of problems follow a common pipeline made by three modules: \emph{perception}, \emph{planning} and \emph{control} modules~\cite{dlr-survey}.
\par
In this extended abstract, we present preliminary results of our offline and online planning strategies for loco-manipulation tasks, applied to legged robots.
The perception module uses state-of-art algorithms to process a dense point cloud generated by the Intel$^{\text{\textregistered}}$ Realsense$^{\text{\texttrademark}}$ D435i depth camera embedded on the robot.
The planner layer is accompanied by a control layer which moves the robot through an impedance controller using the references computed by the planner.
At the current stage, the offline planning strategy has been tested on the real humanoid COMAN+ using a dedicated control layer that executes the planned trajectory once available. 
Recently, we started developing the online planning strategy, which has been successfully tested on the CENTAURO robot executing a wheeled-locomotion task. 
We are currently working towards the design of a unified planning and control framework, which is paramount skill for the success deployment and use of legged robots in challenging environments. 
%
%
\section{Offline Planning}
\label{sec:OfflinePlanning}
\begin{table*}[!t]
\centering
\caption{Averaged performance data of the stance planner.\label{table:PerformanceOffline}}
\begin{tabular}{cccccc}
\hline
\begin{tabular}{cc} Task \end{tabular}
& \begin{tabular}{cc} Planning \\ time (s)\end{tabular}
& \begin{tabular}{cc} Transition \\ generation time (s)\end{tabular}
& \begin{tabular}{cc} Number of \\ iterations  \end{tabular}
& \begin{tabular}{cc} Number of \\ vertices in ${\cal T}$ \end{tabular}
& \begin{tabular}{cc} Number of \\ stances in ${\cal S}_{\sigma}$ \end{tabular} \\  
\hline
Ladder climbing & 43.60 & 37.64 & 1926.14 & 205.34 & 39.04 \\
Parallel walls climbing & 175.37 & 171.01 & 1557.37 & 151.28 & 44.12 \\
Quadrupedal walking & 62.69 & 55.99 & 2540.10 & 228.32 & 53.19 \\
Standing up & 7.56 & 6.02 & 459.24 & 62.19 & 17.00 \\
\hline
\end{tabular}
\end{table*}
In this section, we consider the problem of generating appropriate motions for a torque-controlled humanoid robot that is assigned a multi-contact loco-manipulation task, i.e., a task that requires the robot to move within a \emph{static} and \emph{completely known} environment by repeatedly establishing and breaking multiple, non-coplanar contacts. 

To solve such problem, we propose a offline planning scheme consisting of the two sequential sub-planner described in the following.

The \emph{stance planner} is in charge of finding three sequences
\begin{gather*}
    {\cal S}_{\sigma} = \{ \sigma_0, \dots, \sigma_N \}, \\      
    {\cal S}_{q} = \{ \bfq_0, \dots, \bfq_N \}, \\ 
    {\cal S}_{W} = \{ \bfW_{c,0}, \dots, \bfW_{c,N} \},
\end{gather*}
where ${\cal S}_{\sigma}$ is the sequence of $N+1$ stances leading to the desired final stance $\sigma^{\rm fin}$, i.e., $\sigma_{N} = \sigma^{\rm fin}$, while ${\cal S}_{q}$ and ${\cal S}_{W}$ are sequences of associated transitions \cite{HauserIJRR_2008} and contact wrenches, respectively.
In particular, the generic stance ${\sigma_j \in \cal S}_{\sigma}$ is a set of contacts (each one specifying the pose of a certain end-effector in contact with the environment), transition $\bfq_{j} \in {\cal S}_q$ is a feasible configuration for both $\sigma_{j-1}$ and $\sigma_j$, vector $\bfW_{c,{j}} \in {\cal S}_W$ collects the contact wrenches that guarantee static balance at $\bfq_{j}$ using the contacts specified by $\sigma_{j}$.
To produce the sequences ${\cal S}_{\sigma}$, ${\cal S}_{q}$ and ${\cal S}_{W}$, our stance planner applies a RRT-like strategy; it iteratively constructs a tree $\cal T$ in the search space, where a vertex $v = \langle \sigma, \bfq, \bfW_c \rangle$ consists of a stance $\sigma$, a configuration $\bfq$, and a vector $\bfW_c$ of contact wrenches, while an edge going from vertex $v$ to vertex $v'$ indicates that $\bfq'$ is a transition from $\sigma$ and $\sigma'$.
Transitions are efficiently generated during the planning process using the method presented in \cite{RossiniFrontiers_2021}.

The \emph{whole-body planner}, for each pair of consecutive configurations $\bfq_{j}$ and $\bfq_{j+1}$ ($j = 0, \dots, N-1$) belonging to ${\cal S}_{q}$, computes a feasible whole-body motion $\bfs_j$, i.e., a configuration space trajectory that connects the two and has duration $\delta_j$.
Then, the result of this planner consists in a sequence of motions
\begin{equation*}
    {\cal S}_{s} = \{ \bfs_0, \dots, \bfs_{N-1} \}.
\end{equation*}
To produce the generic motion $\bfs_j \in {\cal S}_{s}$, our whole-body planner adopts the two-stage approach presented in~\cite{RuPaLaMiTs:20}: first, a path consisting of a sequence ${\cal S}_{q,j}$ of feasible configurations joining $\bfq_{j}$ to $\bfq_{j+1}$ is found using the AtlasRRT* algorithm \cite{JaPo:12}; then, a trajectory $\bfs_j$ interpolating the configurations in ${\cal S}_{q,j}$ is computed, simultaneously determining its duration $\delta_j$, via optimization.

To actually execute the planned motions, we have incorporated the presented offline planning scheme into a complete MCPC framework.
This includes a control layer that is responsible for generating online the torque commands $\bar{\bm\tau}$ for the humanoid actuators to execute the planned sequence ${\cal S}_{s}$ of whole-body motions throughout the planned sequence ${\cal S}_{\sigma}$ of stances.
%
%

\section{Online Planning}
\begin{figure*}
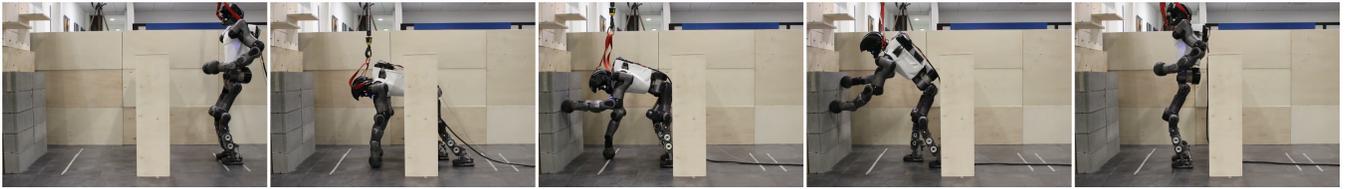

    \centering
    \COMAN
    \caption{COMAN+ sequentially performs the quadrupedal walking and standing up tasks.}
    \label{fig:COMAN}
\end{figure*}
To produce the global configuration space trajectory $\cal{S}_s$, the offline planning scheme previously described assumes that the environment is static.
%
However, a typical working environment for legged robots, is likely to change during the execution of an assigned task, and the robot must be able to rapidly adapt to it to avoid unexpected collisions.
%
To this end, we designed a configuration space local planner based on graph optimization that allows to store the complete trajectory and efficiently perform local modifications whenever needed.
Indeed, the online planner possibly refines the motion $\bfs_j$ to be executed, instead of replanning the global motion $\mathcal{S}_s$, avoiding useless optimizations of future motions that are likely to be affected by further changes of the environment.
Additionally, the solver avoids to get stuck in local minima being guided by the global information coming from the offline planner.
Using the same constraints both for the offline and online planner, they will generate whole-body trajectories projected on the same manifold, thus guaranteeing the output of the offline planner to be feasible also for the online one.
\par
The idea behind the proposed methodology is based on the representation of a global plan as a \emph{hyper graph} where each vertex is a configuration, while each hyper edge represents the error term associated to a generic constraint connecting a certain number of vertices. 
The aim of this optimization is to move the vertices inside the feasible configuration space to minimize the sum of the considered error terms.
\par
We discretize the trajectory $\bfs_j$ into a sequence ${\cal Q}_j$ of equispaced configurations. Then, a vertex is created in correspondence of each configuration $\bfq \in {\cal Q}_j$, while a set of edges is created to build connections between them. In particular, the $k$-th edge will connect a sub-sequence of vertices $\mathcal{Q}_{j,k} \subset \mathcal{Q}_j$.
%
%
Edges involve joint limits, velocity limits, balance, and collision avoidance as well as a cost term to track the global plan.
A continuous error function $\bfe_k({\cal Q}_{j,k})$ is assigned to each edge to penalize those solutions that violate the constraints.
At each iteration, the planner locally refines the global trajectory solving the Non-Linear Least Squared Optimization problem (NLLSO):
\begin{equation}
    \label{eq:cost_function}
    \mathcal{Q}_{j}^* = \operatorname*{argmin}_{\mathcal{Q}_{j}} ~ \bfF(\mathcal{Q}_{j}) 
\end{equation}
where the cost function is
\begin{equation*}
    \bfF(\mathcal{Q}_j) = \sum_{k=0}^m \bfe_k^T(\mathcal{Q}_{j,k})\Omega_k \bfe_k(\mathcal{Q}_{j,k}) = \sum_{k=0}^m \bfF_k(\mathcal{Q}_{j,k})
\end{equation*}
%
with $\Omega_k$ being a diagonal square weight matrix having the same number of rows and columns as the associated error function $\bfe_k(\mathcal{Q}_{j,k})$.
A solution to~\eqref{eq:cost_function} can be found using the Levenberg-Marquardt algorithm which finds the optimal step size $\Delta \mathcal{Q}_j^*$
\begin{equation}
    (\bfH + \lambda\bfI)\Delta \mathcal{Q}^*_j = -\bfb
    \label{eq:update}
\end{equation}
with $\lambda$ being a damping factor used to control the step size $\Delta \mathcal{Q}_j^*$ on non-linear surfaces, $\bfb = \sum_{k=0}^m \bfe_k^T\Omega_k \bfJ_k$, and $\bfH = \sum_{k=0}^m \bfJ_k^T \Omega_k \bfJ_k$ being the $(n \cdot N) \times (n \cdot N)$ information matrix of the system (i.e., Hessian) and is sparse by construction, since the objective functions involved depend on small subsets of neighbours states. $\bfJ_k$ is the $(n \cdot N) \times (n \cdot N)$ Jacobian of $\bfe_k(\mathcal{Q}_j)$ evaluated in $\mathcal{Q}_j$. 
The optimal refined trajectory is then found updating the initial guess recursively:
\begin{equation}
    \mathcal{Q}_j^* = \mathcal{Q}_j + \Delta \mathcal{Q}_j^*.
    \label{eq:optimal_solution}
\end{equation}
The optimized sequence of configurations $\mathcal{Q}_j^*$ is then provided to the robot and used as a new refined reference.
\section{Results}
Both the offline and online planner were tested independently. Results are collected in the following section.
%
%
\subsection{Offline Planning}
We now present some simulation and experimental results obtained using our offline planning scheme on COMAN+, a torque-controlled humanoid robot designed at Istituto Italiano di Tecnologia.

\begin{figure}
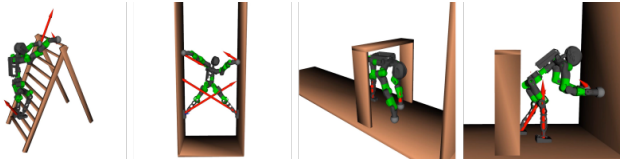

    \centering
    \OfflineTasks
    \caption{The four considered tasks. For each of them, a snapshot from a typical planning solution is provided.}
    \label{fig:MultiContactScenarios}
\end{figure}
To analyze the performance of our offline planning scheme, we performed a simulation campaign on an Intel Core i7-7500U CPU running at 2.70 GHz.
We considered four different multi-contact loco-manipulation tasks (Fig.~\ref{fig:MultiContactScenarios}) described in the following.
\begin{enumerate}
    \item \textit{Ladder climbing}. The robot must climb a ladder that is located in front of it.
    \item \textit{Parallel walls climbing}. The robot must climb vertically between two parallel walls located on its left and right flanks.
    \item \textit{Quadrupedal walking}. The robot must go through a narrow passage that can not be navigated by standard bipedal walking.
    \item \textit{Standing up}. The robot must stand upright, possibly exploiting a wall located in front of it as support.
\end{enumerate}
Since our planning scheme is randomized (as both the stance and whole-body planners rely on probabilistic strategies), we performed 100 runs for each of the four tasks.
%
Table~\ref{table:PerformanceOffline} collects the most significant performance data, averaged over the $100$ runs, of the proposed stance planner. For each task, we report the time needed by the stance planner to find a solution, the time spent in generating transitions, the number of performed iterations, the number of vertices in the constructed tree $\cal T$, and the number of stances in the solution sequence ${\cal S}_\sigma$. 
To show the feasibility of the planned motions on the real robotic platform, we performed a complete experiment in which COMAN+ was requested to sequentially perform the quadrupedal walking and standing up tasks in a purposely constructed scenario (Fig.\ref{fig:COMAN}) \footnote{\url{https://youtu.be/zS44CegGqow}}.
\subsection{Online Planner}
The proposed perception-based online planner was evaluated through a number of simulation studies and experimental trials carried out on the CENTAURO robot~\cite{CENTAURO} reduced considering the joints of the lower body as the only active joints, resulting in a 24 DoFs robot.
The robot is equipped with two Intel$^{\text{\textregistered}}$ Realsense$^{\text{\texttrademark}}$ D435i depth cameras mounted on its pelvis and head.
No other external sensor has been used to perceive the environment.
\par
\begin{figure}
  \begin{subfigure}[]{0.23\columnwidth}
    \centering
    \includegraphics[width=\columnwidth]{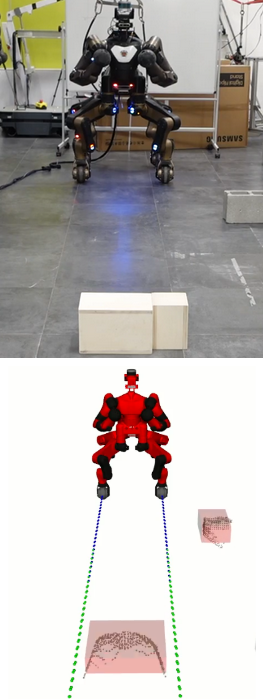}
    \caption{} 
    \label{subfig:object}
  \end{subfigure}
  \hfill
  \begin{subfigure}[]{0.23\columnwidth}
    \centering
    \includegraphics[width=\columnwidth]{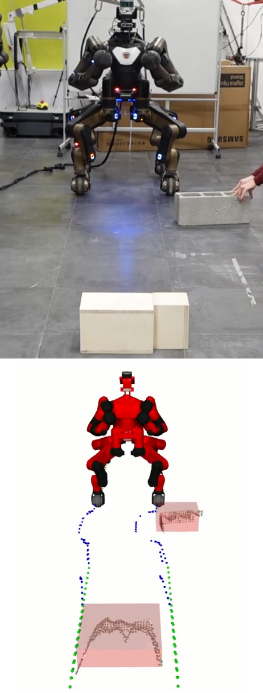}
    \caption{}
    \label{subfig:first_obstacle_front}
  \end{subfigure}
   \hfill
  \begin{subfigure}[]{0.23\columnwidth}
    \centering
    \includegraphics[width=\columnwidth]{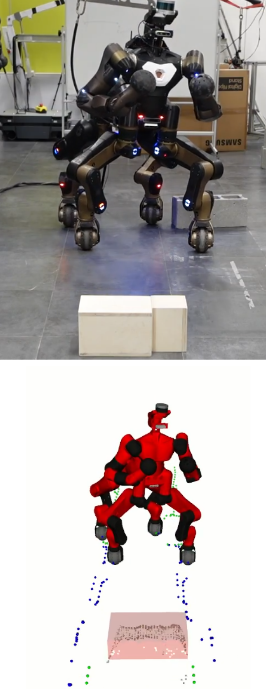}
    \caption{}
    \label{subfig:first_obstacle_rear}
  \end{subfigure}
  \hfill
    \begin{subfigure}[]{0.23\columnwidth}
    \centering
    \includegraphics[width=\columnwidth]{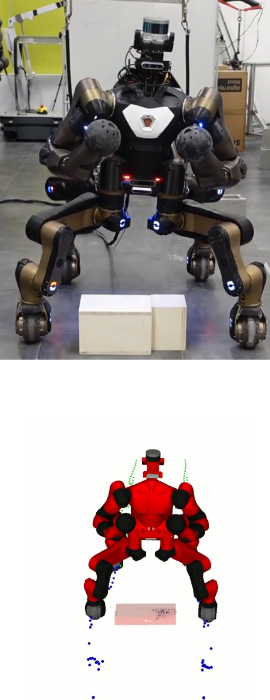}
    \caption{}
    \label{subfig:second_obstacle}
  \end{subfigure}
  \hfill
  \caption{Screenshots from the experiment carried out with the CENTAURO robot. Blue an green dots are the poses of the four wheels corresponding to vertices inside and outside the moving horizon of the planner.}
  \label{fig:exp}
\end{figure}

 
The online planner is tested independently from the offline global planner and a straight wheeled motion is used as the initial global plan, moving the robot 4m forward.
In this scenario, the stance does not change and the robot moves simply rolling and steering its wheels instead of taking steps.
The resulting trajectory is discretized into 50 configurations and the online planner works locally with a moving horizon of 20 vertices, thus ignoring the furthest ones (Fig.\ref{subfig:object}).
However, the global trajectory can be arbitrarily long, including a higher number of vertices without under performing the local planner, which depends on the number of local vertices involved in the refinement only.
During the execution of the task, obstacles are dynamically introduced to force the robot to refine the whole-body trajectory, successfully avoiding any collision (Figs.~\ref{subfig:first_obstacle_front}-\ref{subfig:first_obstacle_rear}-\ref{subfig:second_obstacle}) \footnote{\url{https://youtu.be/qaBNzVrZtF4}}.
%
%
%
%
%
\par
The computation time fluctuates around 0.07s during the execution of the trajectory.
%
%
%
%
\section{Conclusion}
This extended abstract presents our recent work on planning strategies to allow autonomous motion generation of legged robots in cluttered environments. 
Specifically, our planners explore the feasible configuration space. 
The offline planner finds a feasible global plan, that can be used as an initial guess for the online planner, which refines the whole-body trajectory whenever required.
\par
Experiments and simulations were carried out on the two planning modules separately, using two different robotic platforms, i.e., the COMAN+ humanoid and the CENTAURO robot. 
Preliminary results confirmed the possibility to use such a offline and online planning strategy to generate and adjust a feasible configuration space trajectory in general dynamic environments. 
%
%
Future work will focus on merging the two presented offline and online planners into a unified framework allowing the real-time generation of appropriate motions to accomplish loco-manipulation tasks with legged robots.
%
\bibliographystyle{IEEEtran}
\bibliography{PlanningWorkshop}

\end{document}